\documentclass[10pt,twocolumn,letterpaper]{article}

\newif\ifarxiv \arxivtrue

\ifarxiv
    \usepackage[pagenumbers]{iccv} %
\else
    \usepackage[review]{iccv}      %
\fi

\usepackage{amsmath}

\usepackage{dsfont}
\usepackage{amsthm}
\usepackage{overpic}
\usepackage{contour}
\contourlength{1pt}
\usepackage{xspace}
\usepackage{sidecap}
\usepackage{rotating}
\usepackage{booktabs} %
\usepackage{caption}
\usepackage{subcaption}
\usepackage{microtype}
\usepackage{booktabs}
\usepackage{enumitem}
\usepackage{amsthm}
\usepackage{multirow}
\usepackage{tabularx}
\usepackage{nicefrac}
\usepackage{algorithm}
\usepackage{algorithmicx}
\usepackage{algpseudocode}
\usepackage{chngcntr}
\usepackage{apptools}
\usepackage[super]{nth}
\usepackage{epigraph}
\usepackage{makecell}
\usepackage{amsmath}
\usepackage{amsfonts }

\usepackage{ tipa }
\usepackage{adjustbox }
\usepackage[dvipsnames]{xcolor}

\AtAppendix{\counterwithin{lemma}{section}}

\newcommand{\LL}{\mathcal{L}}

\newcommand{\EE}{\mathcal{E}}

\newlength{\indentlaenge}
\newlength{\mylength}
\newlength{\mylengthzwei}
\def\cput(#1,#2)#3{\put(#1,#2){\hbox to 0pt{\hss{#3}\hss}}}
\def\lput(#1,#2)#3{\put(#1,#2){\hbox to 0pt{\hss{#3}}}}
\def\rput(#1,#2)#3{\put(#1,#2){\hbox to 0pt{{#3}\hss}}}

\newcommand{\method}{\textit{LoRA-MDM}\xspace}

\definecolor{Blue9}{rgb}{0.098,0.3,0.9}
\definecolor{Orange6}{rgb}{0.992, 0.494, 0.078}
\definecolor{Green6}{rgb}{0.251, 0.753, 0.341}
\definecolor{Red6}{rgb}{0.98, 0.322, 0.322}
\definecolor{Purple}{rgb}{0.896, 0.39, 1.0}

\definecolor{iccvblue}{rgb}{0.21,0.49,0.74}
\usepackage[pagebackref,breaklinks,colorlinks,allcolors=iccvblue]{hyperref}

\title{
\ifarxiv
    Dance Like a Chicken:\\ Low-Rank Stylization for Human Motion Diffusion
\else
    Dance Like a Chicken: Low-Rank Stylization for Human Motion Diffusion
\fi
}

\author{Haim Sawdayee \\
        Tel Aviv University \\
        \tt\small haimsawdayee@mail.tau.ac.il
        \and
        Chuan Guo\\
        Snap Inc.\\
        \and
        Guy Tevet\\
        Tel Aviv University\\
        \and 
        Bing Zhou\\
        Snap Inc.\\
        \and
        Jian Wang\\
        Snap Inc.\\
        \and
        Amit H. Bermano\\
        Tel Aviv University
    }

\begin{document}

\twocolumn[{%
\renewcommand\twocolumn[1][]{#1}%
\vspace{-40pt}
\maketitle
\vspace{-25pt}
\includegraphics[width=\textwidth]{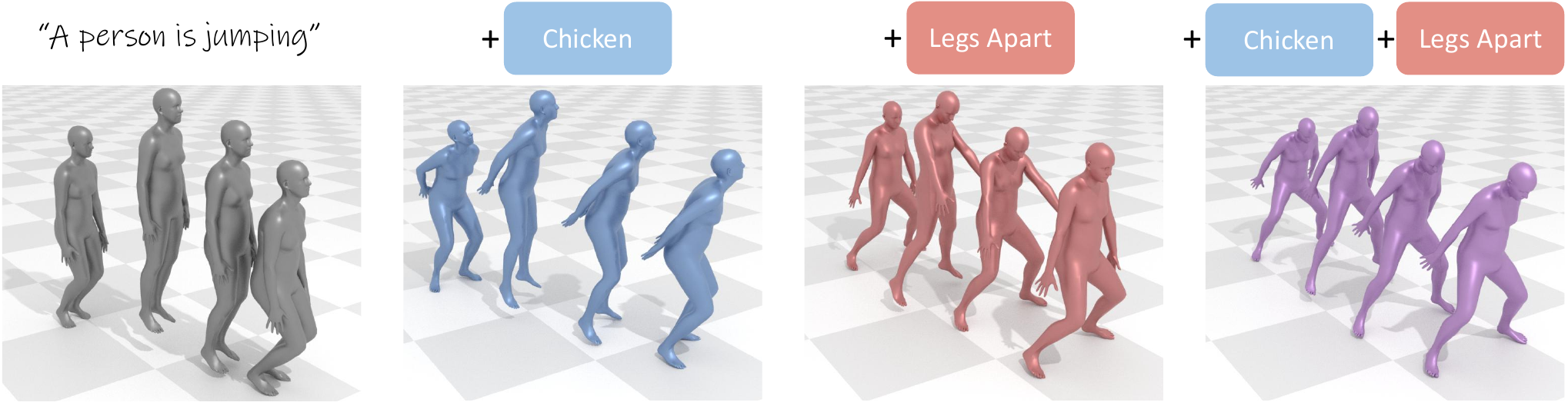}
\captionof{figure}{\method generates stylized motions using Low-Rank Adaptation of the Motion Diffusion Model (MDM).
Left: motions generated using the base model. 
Middle: stylized generation using \method. 
Right: Style mixing. 
\vspace{1em}
}
\label{fig:teaser}
}]

\begin{abstract}

    Text-to-motion generative models span a wide range of 3D human actions but struggle with nuanced stylistic attributes such as a "Chicken" style.
    Due to the scarcity of style-specific data,
    existing approaches 
    pull the generative prior towards 
    a reference style, 
    which often results in out-of-distribution low quality generations.
    In this work, we introduce \method, 
    a lightweight framework for motion stylization that generalizes to complex actions while maintaining editability.
    Our key insight is that adapting the generative prior to include the style, while preserving its overall distribution, is more effective than modifying each individual motion during generation.
    Building on this idea,
    \method learns to adapt the prior to include the reference style using only a few samples. 
    The style can then be used in the context of different textual prompts for generation. 
    The low-rank adaptation shifts the motion manifold in a semantically meaningful way, enabling realistic style infusion even for actions not present in the reference samples.
    Moreover, preserving the distribution structure enables advanced operations such as style blending and motion editing.
    We compare \method to state-of-the-art stylized motion generation methods and demonstrate a favorable balance between text fidelity and style consistency.
    \ifarxiv
    Project page at \url{https://haimsaw.github.io/LoRA-MDM/}
    \fi 

\end{abstract}

\section{Introduction}
\label{sec:intro}

Recent advancements in human motion generation conditioned on text have demonstrated impressive results, producing diverse and high-fidelity motions closely aligned with textual descriptions \cite{tevet2023human,dabral2023mofusion,zhang2023generating}. 
However, these models struggle to convey nuanced attributes, including profiles such as that of an elderly person, and fail to interpret abstract descriptions, such as "a Superman style". 
This limitation stems from the nature of existing text-to-motion datasets \cite{BABEL:CVPR:2021,plappert2016kit,Guo_2022_CVPR}, which lack rich representations of styles and subtle nuances. In contrast, style-specific datasets \cite{aberman2020unpaired,mason2022real} provide better coverage of these nuances, but they are limited to a small and simple set of motions per style, falling short of capturing the wide variety of human actions.

Using the latter datasets, a handful of approaches tackle motion stylization, or infusing a given style into the generation process of another motion. 
Specifically the two leading approaches either inject attention features from a given style motion into the generation process \cite{raab2024monkey}, or combine a ControlNet~\cite{zhang2023adding} with a classifier trained over the styles to perform guidance during the generation process \cite{zhong2025smoodi}. In both cases, the generative process, based on a diffusion prior, is pushed out of its distribution to match the desired style, yielding lower overall quality, generalization, and editability.

In this paper we introduce \method, a simple and efficient stylized motion generation method, yielding quality and editability for diverse stylized actions. Drawing inspiration from image editing and, specifically, personalization literature \cite{gal2022stylegan,ruiz2023dreambooth}, our main insight is that augmenting the generation process with a given style means venturing into regions unfamiliar to the generative prior. Instead, it is better to shift the generative prior itself to include the style, while preserving its overall distribution structure. 

This insight motivates the use of Low-Rank Adaptation (LoRA) \cite{hu2021lora}. LoRA augments the weights of the prior in a low-ranked manner, which can be intuitively thought of as low frequency deformations to the target distribution manifold.
Using LoRA, \method learns to embed a style within the output domain of the model, which can then be used to synthesize stylized motions in the context of different textual prompts. As we demonstrate, this offers two advantages: First, low-ranked augmentation diminishes forgetting, which in turn increases generation quality for stylized actions not seen during training.
For example, directly fine-tuning a generated walking motion to walk like a chicken has little (or negative) influence on how the prior will generate kicks. Instead, our low-ranked tuning and structure preservation loss encourage the model to also kick like a chicken, since the changes affect the entire manifold. 
Second, because the generated stylized motions are in-distribution for the model, advanced operations such as stylized motion editing and style blending are supported, in sharp contrast to traditional motion style-transfer method.

Using a pretrained state-of-the-art motion diffusion generative model and a LoRA trained on one or more styles, we demonstrate \method produces superior results compared to leading motion stylization approaches, qualitatively, quantitatively, and through a user study, both in terms of motion quality, and adherence to style or styles. Since the prior's capabilities are not hampered, we also demonstrate that \method naturally combines with existing motion editing techniques. Specifically, we demonstrate trajectory control capabilities for generated stylized actions. 
In the future, we hope this personalization take on motion style can help set common practices for applying LoRA to other motion domains, and will inspire extending motion generation capabilities to other underrepresented axes.

\section{Related Work}
\label{sec:related}

\subsection{Human Motion Synthesis} 

\paragraph{Multimodal generation.} In recent year, human motion generation has experienced significant progress with the advent of Transformers~\cite{vaswani2017attention} and deep generative models~\cite{ho2020denoising,goodfellow2020generative,kingma2013auto}. Motions can now be generated from a variety of input modalities, including action categories~\cite{petrovich2021action,guo2020action2motion,xu2023actformer,lucas2022posegpt}, music signals~\cite{alexanderson2023listen,siyao2022bailando,lee2019dancing,tseng2023edge}, and textual descriptions~\cite{tevet2023human,Guo_2022_CVPR,tevet2022motionclip,zhang2022motiondiffuse,guo2024momask,guo2022tm2t,shafir2024human,chen2023executing}. Early works achieved plausible 3D motion generation by leveraging variational autoencoders (VAEs) equipped with temporal models such as RNNs~\cite{Guo_2022_CVPR, guo2020action2motion, lee2019dancing} and Transformers~\cite{petrovich2021action, petrovich2022temos}. However, these models often struggled to learn complex actions and handle fine-grained conditionality. To address this, a subsequent line of research modeled human motion generation in a discrete space. These approaches utilized VQ-VAE~\cite{van2017neural} to map human motions into discrete motion tokens, which were then generated from various modalities using language models like GPTs~\cite{zhang2023generating, guo2022tm2t, siyao2022bailando, lucas2022posegpt} or generative masked Transformers~\cite{guo2024momask, pinyoanuntapong2024mmm}. Diffusion models have also emerged as a powerful tool for motion synthesis due to their exceptional expressiveness. MDM~\cite{tevet2023human} and MotionDiffuse~\cite{zhang2022motiondiffuse} applied diffusion models directly in the raw motion feature space, while techniques like latent diffusion~\cite{chen2023executing} and latent consistency models~\cite{dai2025motionlcm} have been adopted to enhance the efficiency of diffusion-based human motion synthesis. 

\paragraph{Motion style transfer.} Motion style transfer aims to reenact an existing motion while incorporating the style characteristics specified by a reference style motion or label. Traditional methods, such as motion matching~\cite{xia2015realtime} and gram matrix optimization~\cite{holden2017fast, du2019stylistic}, are limited in scalability and efficiency. Recent advances in data-driven deep learning models have enabled more efficient and faithful motion style transfer. Abermann et al.~\cite{aberman2020unpaired} introduced the use of AdaIN to disentangle style and content features, which has since been adapted in works such as Guo et al.\cite{guo2024generative}, MotionPuzzle~\cite{jang2022motion}, Tang et al.\cite{tang2024decoupling}, and MOCHA\cite{jang2023mocha}. Adversarial learning has also been widely employed to enhance style-awareness~\cite{aberman2020unpaired, tao2022style}. More recently, studies like MoMo~\cite{raab2024monkey} and MotionCLR~\cite{chen2024motionclr} suggest that pre-trained motion diffusion models can achieve zero-shot style transfer by manipulating attention layers. Additionally, Li et al.~\cite{li2024walkthedog} proposed a periodic autoencoder to extract phase features from motion, demonstrating style transfer as a byproduct of switching periodic phase features. 
More recently, MulSMo\cite{li2024mulsmo} suggested style mapping using a bidirectional control flow.

\paragraph{Stylized motion generation.} While style transfer focuses on re-enacting existing motions, our work aims to develop generative models capable of directly producing motions of specific styles given textual descriptions. To the best of our knowledge, SMooDi~\cite{zhong2025smoodi} is the only prior work in this area. SMooDi leveraged a pre-trained motion-to-text model~\cite{guo2022tm2t} to caption the style dataset, 100STYLE~\cite{mason2022real}, creating new text-motion pairs for style-specific motions. These pairs were then used to learn a style adaptor added on an existing motion diffusion model~\cite{chen2023executing} using techniques from ControlNet~\cite{zhang2023adding} and classifier-guided sampling~\cite{dhariwal2021diffusion}. However, SMooDi suffers from overfitting on the style motion dataset, resulting in lifeless motions when handling out-of-domain prompts.

\begin{figure*}
    \centering
    \includegraphics[width=1\textwidth]{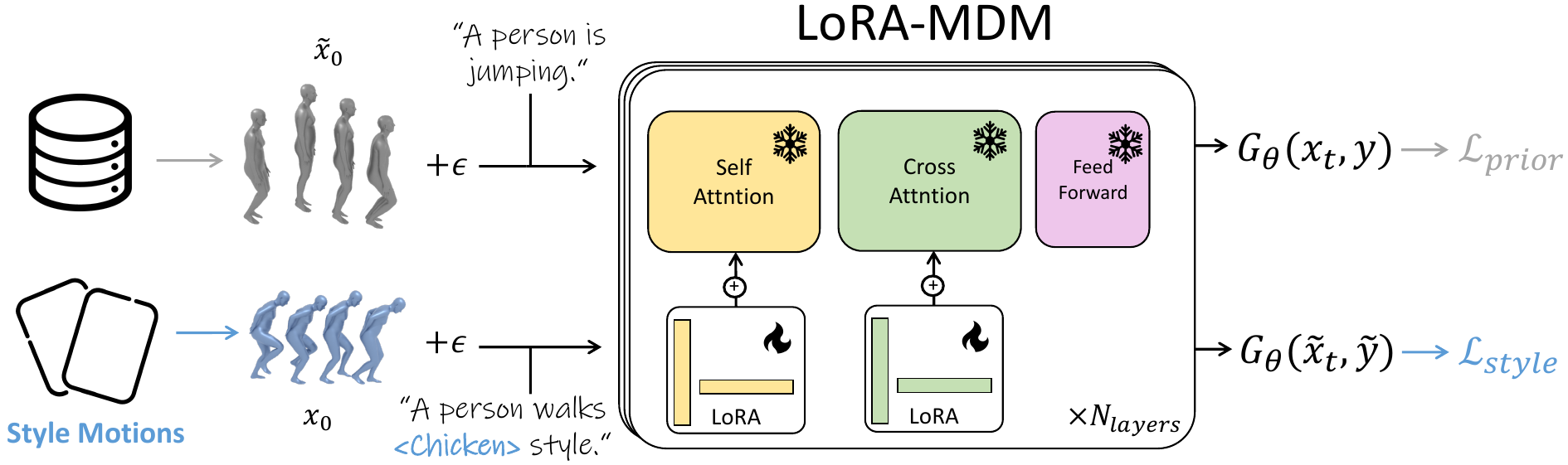}
    \caption{\textbf{\method Overview.} 
    We train Low-Rank Adaptations for the attention layers of the Motion Diffusion Model (MDM) \cite{tevet2023human} to shift it toward a given style (e.g., "Chicken") while preserving its structure. The style is represented by a handful of reference motions and bound with a special text token, marked $<\cdot>$. The loss terms $\LL_{style}$ and $\LL_{prior}$, both based on the original $\LL_{simple}$, learn the style adaptation and preserve the manifold structure, respectively.}
    \label{fig:overview}
\end{figure*}

\subsection{Personalized Image Generation}
Personalized image generation aims to introduce new visual concepts or styles to base text-to-image (T2I) models using reference images, which is directly relevant to our task. DreamBooth~\cite{ruiz2023dreambooth} fine-tunes the entire network with a preservation loss and requires only a few images. Textual Inversion~\cite{gal2022image} optimizes a token embedding for each new concept. However, both methods involve fine-tuning the entire network, which can be inefficient. Low-Rank Adaptation (LoRA)\cite{hu2021lora} improves fine-tuning efficiency by introducing low-rank layers to the base model and optimizing only the weight residuals. This approach has been widely adopted in personalized T2I models\cite{guo2023animatediff,sohn2023styledrop,kumari2023multi,ryu2023low,tewel2023key,arar2024palp}. Subsequent studies have expanded this personalization approach to handle multiple concepts~\cite{gu2024mix,po2024orthogonal}. AnimateDiff~\cite{guo2023animatediff} leverages existing personalized T2I models, converting them into animation generators by integrating a pre-trained motion module. Some works bypass the model training phase with dedicated personalization encoders~\cite{gal2023encoder,ye2023ip,shi2024instantbooth} or by simply swapping certain attention layers~\cite{jeong2024visual,hertz2024style}. Given the relatively small size of human motion datasets, we adopt LoRA for stylized motion generation.

\section{Method}
\label{sec:method}

\subsection{Preliminaries}

\paragraph{Motion Representation.} 
We adopt the 263-dimensional human motion representation proposed by~\citet{Guo_2022_CVPR}. Each pose in a motion sequence $x\in \mathbb{R}^{N\times 263}$ encodes root kinematics (angular velocity, linear velocities, and height), joint-level information (positions, velocities, and rotations in root space), and binary indicators for foot-ground contacts.
\ifarxiv
    See Appendix~\ref{subsec:sup_repr} for full details.
\fi

\paragraph{Denoising Diffusion.}
We follow the common practice in motion diffusion generation and use the Denoising Diffusion Probabilistic Models (DDPM) formulation~\cite{ho2020denoising}.
The diffusion process consists of forward and backward processes. During training, in the forward pass, we add noise $\epsilon$ to the clean data $x_0=x$ according to:
\begin{equation*}
\label{eq:diffusion}
    x_t = \sqrt{\bar{\alpha}_t}x_0 + \sqrt{1-\bar{\alpha}_t}\epsilon,\ \ \ \epsilon \sim \mathcal{N}(0,I)
\end{equation*}
where $\alpha_t \in (0, 1)$ are predefined constants, $t$ is the diffusion timestep, and $t \in [0, T]$. The training objective of the generator $G$ is to predict the $x_0$ from the noisy observations $x_t$ and condition $c$ as:

\begin{equation*}
    \mathcal{L}_{simple} = \mathbb{E}_{x_0\sim q(x_0|c), t\sim[1, T]}
    \left[\lVert x_0 - G(x_t, t, c) \rVert_2^2 \right]
\end{equation*}
During inference, $G(x_t, t, \mathbf{c})$ is iteratively applied from $t=T$ down to $t=0$ to recover $\hat{\mathbf{x}}_0$. 

\paragraph{Motion Diffusion Model (MDM).}
MDM~\cite{tevet2023human} is a popular diffusion model for human motion, which was proven as a good baseline for many follow-up works~\cite{raab2023single,karunratanakul2023guided,chen2024taming,cohan2024flexible,tevet2024closd,kapon2024mas}. In this work we use its transformer-decoder version with 
DistilBERT~\cite{sanh2019distilbert} text encoding as implemented in \cite{tevet2022motiondiffusion}.

\paragraph{Low-Rank Adaptation (LoRA).}
Given a pretrained weight matrix $W_0 \in \mathbb{R}^{d \times k}$, traditional fine-tuning updates $W_0$ directly to a new weight matrix $W = W_0 + \Delta W$. However, LoRA~\cite{hu2021lora} constrains the update matrix $\Delta W$ to be of low rank:
\begin{equation}
    \Delta W = AB, \ A \in \mathbb{R}^{d \times r}, \ B \in \mathbb{R}^{r \times k}, \ r \ll \min(d, k).
\end{equation}
Here, $r$ is the rank of the adaptation, which is typically much smaller than $d$ and $k$, enabling significant parameter reduction. During training, only $A$ and $B$ are optimized while $W_0$ remains frozen. %

\subsection{LoRA-MDM}

The method's overview is illustrated in Figure~\ref{fig:overview}.
Our goal is to generate high-quality human motions in a specified style while preserving the diversity and generalization ability of a pre-trained motion diffusion model (MDM).
Conditioning motion generation on style-specific data often leads to a degradation in motion quality and generalization due to the limited diversity of style datasets.
On the other hand, fine-tuning the whole model using the style data, which includes a handful of motions per style, may lead to an overfit and the forgetting of underrepresented actions.
Inspired by personalization literature~\cite{gal2022stylegan,ruiz2023dreambooth}, we propose a lightweight adaptation method based on Low-Rank Adaptation (LoRA), which enables the model to internalize new styles while maintaining its learned motion manifold.

To incorporate a motion style, $s$, into the generative process, we introduce trainable LoRA weights $\theta$ into the self- and cross-attention projection matrices of the MDM denoiser $G$, while keeping the rest of the MDM weights frozen.
Additionally, we initialize a new text token $\langle s \rangle$ to represent the target style. The adapted model is denoted as $G_\theta$.  

As shown in the top part of Figure~\ref{fig:overview}, we embed a given style $s$ into the prior's motion manifold by training the LoRA parameters using a style reconstruction loss,

\begin{equation*}
    \LL_{style} = E_{x_0, y, t}\left[\left|\left| G_\theta \left(x_{t}, y+\textrm{\textit{"in $\langle s \rangle$ style"}}, t \right) - x_0  \right|\right|^2 \right].
    \label{eq:l_style}
\end{equation*}

Given a stylized motion $x_0$, a short text description $y$, and a noise level $t$, we generate a noised version $x_t$ and denoise conditioned on both $y$ and the style token $\langle s \rangle$ (e.g., \textit{"A person [...] \textbf{in $\langle s \rangle$ style}"}).

 A major challenge in training style-aware motion generation models is the risk of distorting the motion manifold, leading to degraded motion quality, as observed in prior works on personalized images \cite{ruiz2023dreambooth}. 
 The low-rank nature of our approach inherently mitigates this distortion. To diminish the phenomenon even further, we introduce a \textit{Prior Preservation Loss}: an additional reconstruction loss,
 
 \begin{equation*}
    \LL_{prior} = E_{\tilde x_0, y, t}\left[ \left|\left| G_\theta \left(\tilde x_{t}, y, t \right) - \tilde x_0  \right|\right|^2 \right],
    \label{eq:l_prior}
\end{equation*}
 
applied to non-stylized motions $\tilde{x}_0$ sampled from the larger text-to-motion dataset, on which the model was originally trained. As illustrated in Figure~\ref{fig:overview}, this loss ensures that the generative prior retains its ability to produce high-quality, diverse motions across various actions while being adapted to style data.

The full training objective,

\begin{equation*}
    \LL = \LL_{style} + \lambda \LL_{prior},
    \label{eq:loss}
\end{equation*}

establishes an association between the token $\langle s \rangle$ and the style $s$, enabling the generation of novel motions in the learned style while preserving the structure of the original diffusion model.
By integrating both style embedding and prior preservation, our method effectively captures stylistic nuances while maintaining the ability of the generative model to produce diverse and high-quality motions.

\section{Experiments}

\label{sec:exp}

\begin{figure}[ht!]
    \centering
    \includegraphics[width=1.0\columnwidth]{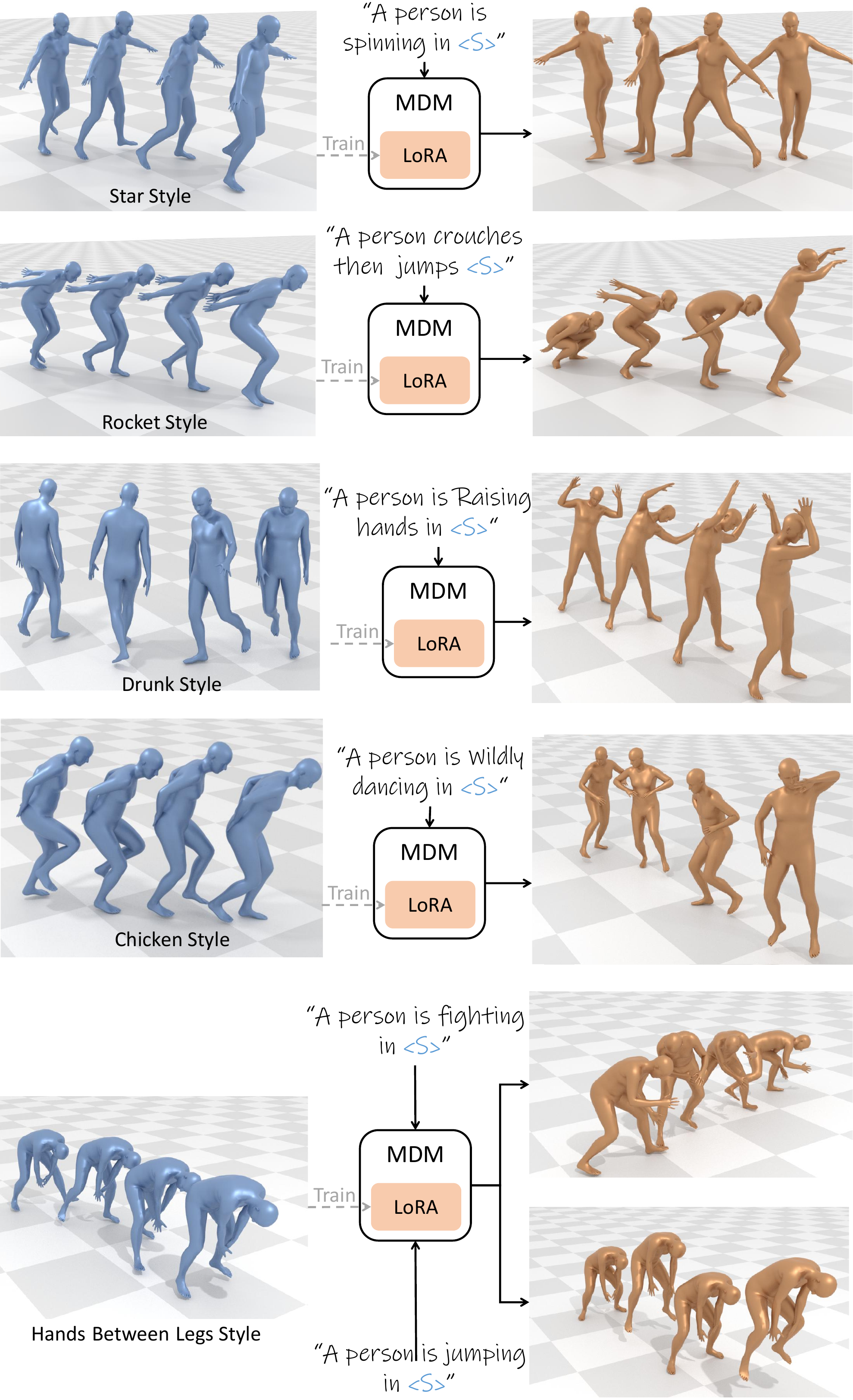}
    \caption{\textbf{Qualitative Results.} The LoRA adapter is trained on a handful of style reference motions (blue, left). Generated motions (orange, right) adhere to both the text prompt and style.
    Notably, the dynamic style “drunk” applied to “raise hands” reveals semantic style transfer, and the “chicken” style with “dance wildly” yields smooth and coherent motion, demonstrating non-trivial generalization.
    Also, the prior fits Capoeira moves to "fighting" and frog leaps to "jumping" to better fit both the text prompt and the learned style.}
    \label{fig:parade}
\end{figure}

\subsection{Evaluation Setting}

\begin{figure*}[t!]
    \centering
    \includegraphics[width=\textwidth]{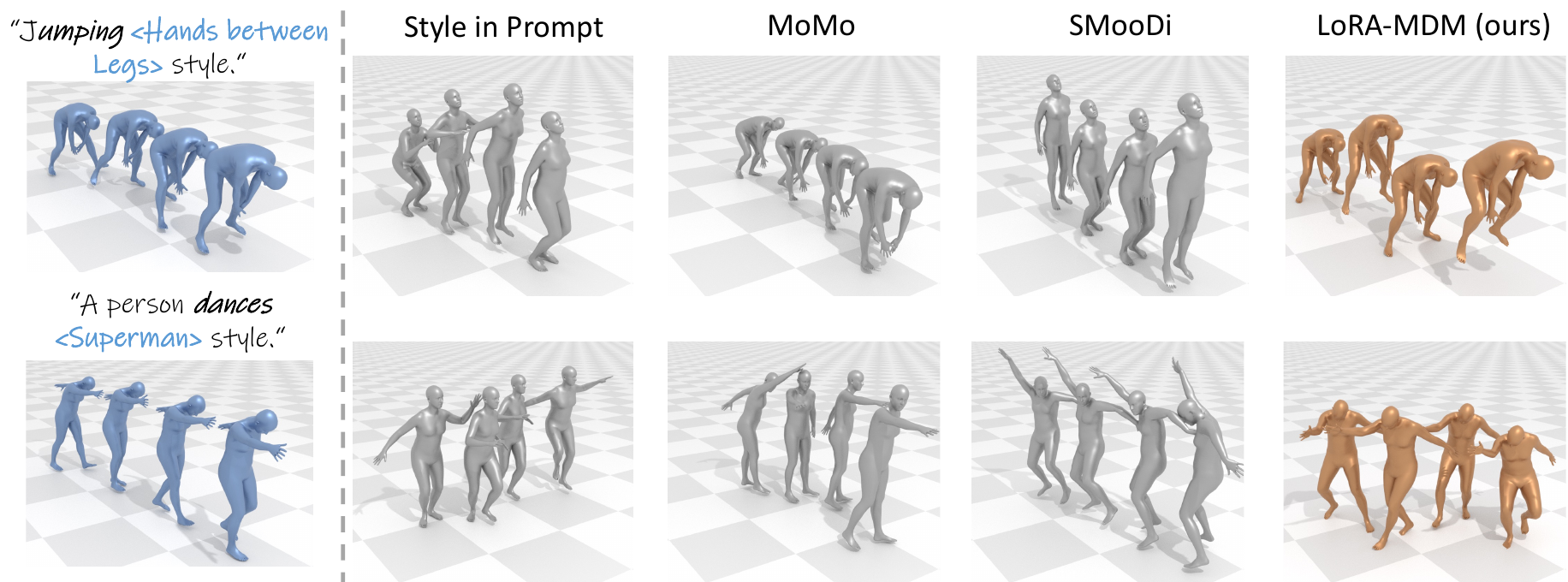}
    \caption{
    \textbf{Comparisons.}
    We compare our generations (orange, right) given a few style references (blue, left) and a text prompt to three baselines (gray, middle). Describing the style in the prompt fails to capture the style nuances; MoMo~\cite{raab2024monkey} nicely fits the style, but fails adhering to the prompt; with SmooDi~\cite{zhong2025smoodi}, results sometimes adhere too strongly to the style, losing plausibility (bottom), and sometimes omit the style to align to the prompt content (top). 
    } 
    \label{fig:visual_comparison}
\end{figure*}

\label{subsec:evaluation}

To evaluate \method, we conducted both a user study and a statistical evaluation.  
Following SMooDi~\cite{zhong2025smoodi}, we assess all baselines across three key dimensions: style alignment, content preservation, and motion quality.  

\paragraph{Implementation Details.}
Our base MDM~\cite{tevet2023human} model uses an 8-layer transformer decoder with latent dimension $512$, where the text is encoded using DistilBERT~\cite{sanh2019distilbert} and injected through the cross-attention layers. The model was trained with $T=100$ for $500K$ training steps with batch size $64$ on a single A100 GPU. We used the implementation in \cite{tevet2022motiondiffusion}.

We trained LoRA for all the attention matrices in the model with a rank of $5$, i.e., about 2\% of the parameters in the original matrix. We used $\lambda=1$ and fine-tuned for each style with $4K$ training steps on a single L40S GPU, using a batch containing all motions of the given style and a learning rate of $10^{-5}$.
\ifarxiv
    See Appendix~\ref{app:impl} for full details.
\fi

\paragraph{Data.}
Our base model was trained on the popular HumanML3D text-to-motion dataset~\cite{Guo_2022_CVPR}.
Then the LoRA training was done with the 100STYLE dataset \cite{mason2022local} for style learning, and HumanML3D for prior preservation.
For compatibility between the two, we use the 100STYLE dataset \cite{mason2022local} version retargeted to SMPL~\cite{SMPL:2015} skeleton by SMooDi~\cite{zhong2025smoodi}.
\ifarxiv
    The full details are in the Appendix.
\fi

This dataset contains motion capture recordings of seven basic locomotions (e.g., walk and run) performed in 100 different styles.  
For each movement, we created a short text description (e.g., \textit{"A person is walking forward."}) to serve as the prompt.

\paragraph{Baselines}
Our first baseline, called \emph{style in prompt} or \emph{prompting}, is the base MDM model, without LoRA training, which gets the style description in the text prompt (e.g., \textit{"A person [...] \textbf{in chicken style}"}).
We introduce this baseline to 
assess whether the model can naturally interpret stylistic descriptors without explicit adaptation.

Additionally, we compare our method to two state-of-the-art approaches for stylized motion generation: \emph{SMooDi}~\cite{zhong2025smoodi} and \emph{MoMo}~\cite{raab2024monkey}.  
SMooDi leverages ControlNet along with classifier-based and classifier-free guidance to condition the generation process on a reference style motion.
MoMo, on the other hand, 
is a training-free method that
utilizes the attention mechanism of MDM to semantically fuse reference "follower" and "leader" motions into one motion. Those motions can be specified either by a prompt or as a reference motion. The authors demonstrate stylized motion generation by injecting the style reference as the "follower" and the content as the "leader". We follow this framework and perform stylized motion generation using the style motion from 100STYLE as the "follower", and the HumanML3D text prompt as the "leader".

\begin{table*}[t!]
    \centering
    \footnotesize
        \begin{tabular}{lccc|ccccc}
            \toprule
             & \makecell{SRA \\ (All)}  & \makecell{SRA \\ (No action)}  & \makecell{SRA \\ (Character)} $\uparrow$ & \makecell{R-precision \\(Top-3)} $\uparrow$ & FID $\downarrow$  & MM Dist $\downarrow$   & Foot skating $\downarrow$   & Diversity  $\rightarrow$  \\
            \hline
            Real & 90.9\% &  97.7\% &  97.7\% &  79.6\%  & -- &  2.96 &  0.057 & 9.05 \\
            \hline
            \method (ours) & 27.3\% & \underline{43.0\%} & 70.3\% & \underline{79.3\%} & \underline{0.31} & \textbf{3.03} & \underline{0.054} & \underline{8.31} \\
            Prompting & 8.3\% & 12.8\% & 40.6\% & \textbf{79.7\% } & \textbf{0.30} & \underline{3.06} & \textbf{0.045} &  10.02 \\
            SMooDi & \underline{27.5\%} & 40.4\% & \underline{74.9\%} & 57.3\% & 1.37 & 4.52 & 0.115 & \textbf{9.68}  \\
            MoMo & \textbf{68.7\%} & \textbf{83.1\%} & \textbf{98.0\%} & 11.7\% & 19.52 & 8.67 & 0.315 & 6.41 \\
            \bottomrule
    
        \end{tabular}
    \caption{
    \textbf{Quantitative results.}
    We compare \method to three baselines and show it provides a better trade-off between style and prompt alignment. Direct \emph{Prompting} fails to capture the style, whereas MoMo~\cite{raab2024monkey} fails to adhere to the text. SMooDi~\cite{zhong2025smoodi} achieves comparable style alignment, yet \method better adheres to the text (R-precision), preserves the manifold (FID, Matching score, and Diversity), and the motion realism (Foot-skating ratio). Diversity is better when closer to the real data value.
    }
    \label{tab:quantitative}
\end{table*}

\subsection{Quantitative Results}

Following SMooDi~\cite{zhong2025smoodi}, we evaluate the quality of generated motions across three key dimensions: style alignment, content preservation, and motion quality.  

To assess style alignment, measure \emph{Style Recognition Accuracy (SRA)}, specifically top-5 accuracy, which quantifies how accurately the generated motion matches the intended style.
Since certain styles inherently contain content-specific information (e.g., $\langle OnHeels \rangle$), which may interfere with the text prompt, we evaluate the SRA metric across three distinct style groups: (1) all available styles, (2) styles without explicit content-related attributes (as Proposed by Zhong et al. \cite{zhong2025smoodi}), and (3) styles that describe character-based motions (e.g., \textit{Cat, Robot}). 
\ifarxiv
    See the Appendix materials for further details.
\fi

The SRA is measured by a pre-trained style classifier.
We find that SMooDi used the same style classifier for both classifier-based guidance and SRA evaluation.
Instead, we trained a separate CNN classifier to ensure a fair comparison. 
Furthermore, as the SMooDi evaluation script generates motions with an unbalanced style distribution that could skew results, we enforced an even style distribution for a more reliable evaluation.
\ifarxiv
    The implementation details of the classifier are in the Appendix~\ref{app:impl}.
\fi

For content preservation, we employ three metrics:
\emph{Fréchet Inception Distance (FID)} measures the distributional similarity between generated and real motions;
\emph{Multi-Modal Distance (MM Dist)} evaluates how well the generated motion maintains consistency with the text prompts;  
\emph{Motion-Retrieval Precision (R-Precision)} measures how often generated motions are retrieved as top matches when queried with their corresponding textual descriptions.

Finally, for motion quality, we measure the \emph{Foot Skating Ratio} and motion \emph{Diversity}.  
\ifarxiv
    See Appendix for further details.
\fi

The results are presented in Table~\ref{tab:quantitative}.
We observe that \method provides a better trade-off between style and prompt alignment. While specifying the style in the prompt adheres to the text, it fails to capture the style; MoMo~\cite{raab2024monkey}, on the other hand, captures the style but ignores the text. SMooDi~\cite{zhong2025smoodi} achieves comparable style alignment, yet \method better adheres to the text (R-precision), preserves the manifold (FID, Matching score, and Diversity), and the motion realism (Foot-skating ratio).

\subsection{User Study}

We conducted a user study on Amazon Mechanical Turk (AMT) to perceptually evaluate our results against two methods: MoMo~\cite{raab2024monkey} and SMooDi~\cite{zhong2025smoodi}. For each baseline, we randomly sampled 100 generated motions from our and the baseline method, conditioned on the same prompts and style references, across various actions and the 'Character' style group. The animations were presented to users side-by-side along with the reference style and input prompt. Users were asked to indicate their preference between the two generated results regarding three key aspects: \textit{style alignment} with the reference, \textit{content alignment} with the text prompt, and \textit{overall motion quality}.

\begin{figure}
    \centering
    \includegraphics[width=0.5\textwidth]{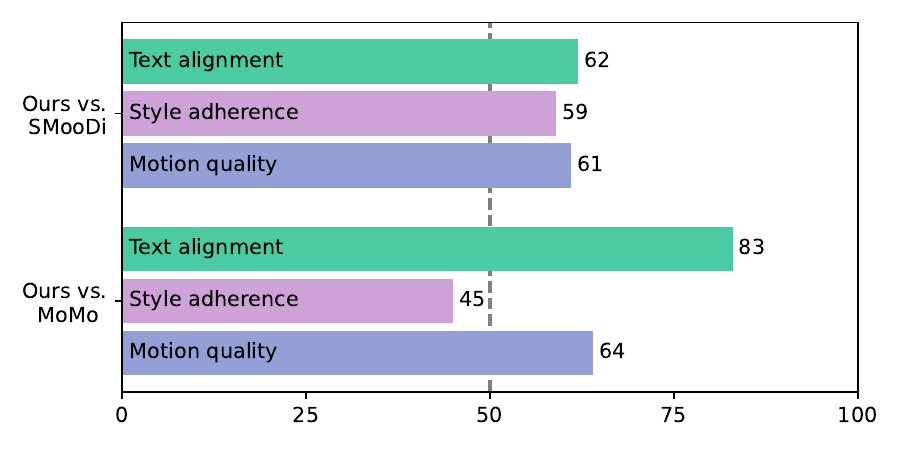}
    \caption{ \textbf{User study}. Subjective comparison of \method to SMooDi~\cite{zhong2025smoodi} and MoMo~\cite{raab2024monkey} in terms of text alignment, style adherence, and overall quality.
    The dashed line marks 50\%. While MoMo is marginally better at style alignment, it fails comparatively to align with the text.}
    \label{fig:user_study}
\end{figure}

The study included 24 distinct users with \textit{master} qualification, each having an approval ratio exceeding 97\% across more than 1000 tasks on AMT. Figure~\ref{fig:user_study} presents the results of this evaluation. Across all assessed aspects, our method consistently achieved approximately 60\% user preference when compared against SMooDi. Aligned with the quantitative evaluation we see here as well that MoMo is marginally better at capturing the style while \method is preferred with respect to text alignment and motion quality.

\subsection{Qualitative Results}
Figure~\ref{fig:parade} 
present motions generated by \method for multiple styles and text prompts.
As illustrated in Figure~\ref{fig:visual_comparison}, \method effectively integrates both style and content into a coherent motion, while other methods struggle to do so.  
When incorporating the style name directly into the prompt, the generated motion adheres to the textual description but mostly fails to capture the intended style, as it is not inherently present in the base model.  
MoMo, while producing motions with prominent style characteristics, struggles to adapt the style to the given prompt due to the misalignment between style and content attention features in the mixed attention mechanism.  
Finally, because of the strong conditioning, the generated motions in SMooDi often drift out of the learned manifold, leading to inconsistent results. As a result, SMooDi sometimes neglects the intended style, while in other cases, it fails to preserve the original content.

\subsection{Ablation Study}
\label{subsec:ablations}

To explore our main design choices, we conducted four ablation studies, all presented in Table~\ref{tab:ablations}.

\begin{figure}
    \centering
    \includegraphics[width=0.5\textwidth]{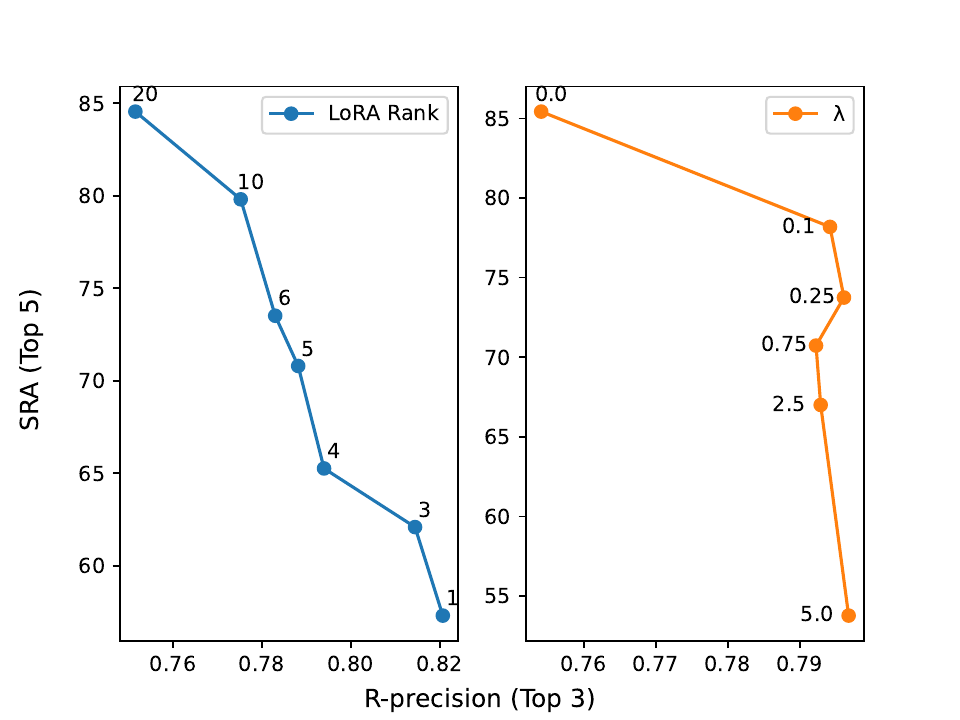}
    \caption{
    \textbf{Hyperparameter trade-offs.}
    We demonstrate the style- and prompt-alignment performance of \method using the SRA and R-precision metrics, respectively (higher is better for both). We get the sweet spot of the two around $\lambda = 0.25$ and $rank = 5$. The full ablation study is in Table~\ref{tab:ablations}.
    }
    \label{fig:ablations}
\end{figure}

\paragraph{LoRA Rank.}
A key feature of LoRA is that restricting the rank also constrains the expressiveness of the adapter.
This trade-off is crucial when adapting a pre-trained model, as it determines the balance between personalization and generalization. If the rank is too low, the style may not be adequately captured, whereas an excessively high rank can lead to overfitting. Figure~\ref{fig:ablations} (Left) illustrates this trade-off and supports our choice of using LoRA.

\paragraph{Loss.}
The personalization vs. generalization trade-off is also affected by the loss balancing factor $\lambda$. Reducing the weight of $\LL_{prior}$ yields motions that better adhere to the style, yet stray from the motion manifold.
Conversely, increasing $\lambda$ improves motion quality but reduces style prominence. (See Figure~\ref{fig:ablations}, Right).

\paragraph{Prior preservation data source.}
We ablated the source distribution of $\LL_{prior}$ by evaluating performance across different data sources: (1) motions generated by the base model with the same actions as in the style dataset, as suggested by Ruiz et al. \cite{ruiz2023dreambooth}; (2) the broader human motion distribution captured in the HumanML3D dataset \cite{Guo_2022_CVPR}; and (3) a combination of both.
As expected, using the HumanML3D dataset improves all metrics except SRA (style prominence).

\paragraph{Adapted weights.}
Since different components of the transformer attention layer influence distinct aspects of the generation process \cite{patashnik2024consolidating, raab2024monkey}, we ablated the effect of adapting the query projection matrices ($W^Q$) and the feedforward layers (FFN). Applying LoRA to the FFN improves SRA but negatively impacts other metrics, whereas adapting $W^Q$ yields comparable overall results. See the bottom part of Table~\ref{tab:ablations} for details.

\begin{table*}[t!]
    \centering
    \footnotesize
    \begin{tabular}{lccccccc}
        \toprule
         & \makecell{SRA \\ (Top-5)} $\uparrow$ &\makecell{R-precision \\ (Top-3) }$\uparrow$ & FID $\downarrow$  & MM Dist $\downarrow$   & Foot skating $\downarrow$   & Diversity  $\rightarrow$  \\
        \midrule

        $\mathrm{rank=3}$ & 62.1\% & \textbf{81.4\%} &  \textbf{0.46} &  \textbf{3.00} &  \textbf{0.039} & \underline{9.36}  \\
        $\mathbf{rank=5}$ & \underline{70.8\%} & \underline{78.8\%} & \underline{0.54} & {3.09} & \underline{0.042} & 10.15 \\
        $\mathrm{rank=10}$ & \textbf{79.8\%} & 77.5\% & 0.94 & 3.25 & 0.046 &  \textbf{9.13} \\

        \hline
        $\lambda = 0$ & \textbf{85.4\%} & 75.4\% & 0.72 & 3.32 & 0.075 & 19.58  \\
        $\boldsymbol{\lambda = 0.25}$  & \underline{73.8\%} &  \textbf{79.6\%} &  \textbf{0.31} &  \textbf{3.02} & \underline{0.054} &  \textbf{10.30 } \\
        $\lambda =0.5$ & 71.6\% & \underline{79.4\%} & \underline{0.38} & \underline{3.04} &  \textbf{0.047} & \underline{10.26} \\
        
        \hline
        Generated & \textbf{79.4\%} & 63.1\% & 2.33 & 4.15 & 0.057 & \textbf{8.66} \\
        \textbf{HumanML3D} & 70.7\% &  \textbf{78.9\%} &  \textbf{0.54} &  \textbf{3.09} &  \textbf{0.042} &  10.14 \\
        Mixed & \underline{76.4\%} & \underline{75.6\%} & \underline{1.24} & \underline{3.37} & \underline{0.034} & \underline{9.60} \\

        \hline
        \textbf{\textit{w $W^Q$, w/o FFN}}  & 70.8\% &  \textbf{79.0\%} & \underline{0.54} & \underline{3.09} & \underline{0.042} &  \underline{10.12 } \\ %
        \textit{w/o $W^Q$, w/o FFN} & 69.2\% & \underline{78.6\%} &  \textbf{0.49} &  \textbf{3.05} &  \textbf{0.041} & \textbf{ 9.55 } \\ %
        \textit{w $W^Q$, w FFN}  & \textbf{95.5\%} & 16.7\% & 9.61 & 7.49 & 0.087 & 7.08 \\ %

        \textit{w/o $W^Q$, w FFN} & \underline{93.5\%} & 17.9\% & 9.33 & 7.31 & 0.082 &  6.13 \\  %
        \bottomrule
    \end{tabular}
    \caption{
    \textbf{Ablation study.}
    We ablate the rank of the LoRA, the loss balancing ($\lambda$), the data source for prior preservation, and the weights on which the LoRA is applied. The trade-off curves of the rank and the $\lambda$ are in Figure~\ref{fig:ablations}. Diversity is better when closer to the real data value. Our selected parameters are highlighted in bold.
    \ifarxiv
        See the full table in Appendix~\ref{subsec:sup_ablations}.
    \fi
    }
    \label{tab:ablations}
\end{table*}

\begin{figure}
    \centering
    \includegraphics[width=0.5\textwidth]{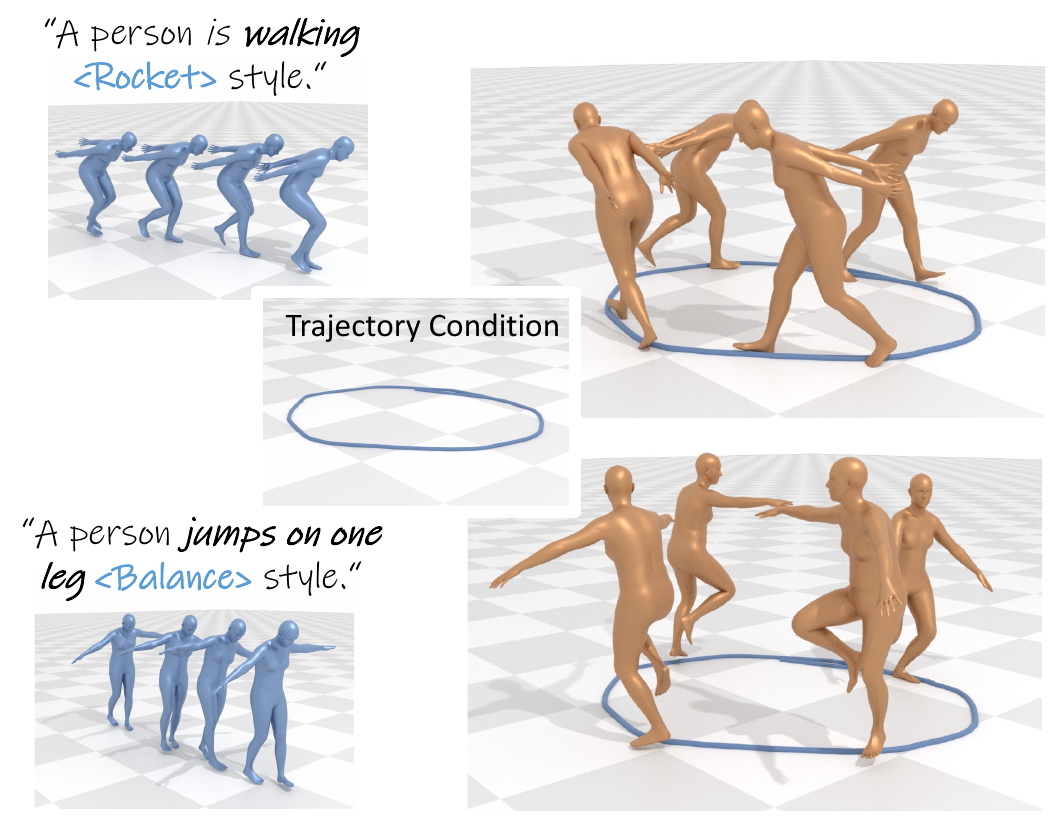}
    \caption{
    \textbf{Trajectory Control.} 
    An important advantage of the manifold preservation is maintaining compatibility with the motion diffusion ecosystem. We demonstrate trajectory control (presented in PriorMDM~\cite{shafir2024human}), integrated with \method.
    }
    \label{fig:edit}
\end{figure}

\section{Applications}
\label{sec:app}

\subsection{Motion Control}
\label{subsec:edit}
One of the most important properties of in-domain generation is editability: While generative models struggle with modifying specific semantic aspects while preserving others, they are still capable of achieving this in a disentangled manner.
For regions that a model is not well familiar with, this is a nearly impossible task. 
Modifying one aspect of motion often requires subtle adjustments to others to maintain realism.
For example, when jumping on one leg, a turn in the trajectory should induce shifting the body's center of mass. 

To demonstrate in-distribution generation, we integrate \method with an off-the-shelf motion editing technique.
We employ the method of \citet{shafir2024human}, which demonstrates that new controls can be added to the MDM backbone through simple self-supervised fine-tuning.
 As can be seen in Figure~\ref{fig:edit}, the control-enhanced, fine-tuned version of MDM naturally integrates with our LoRA-based fine-tuning approach, yielding high-quality stylized motions that are also trajectory-controlled. Specifically, note in Figure~\ref{fig:edit} how the body realistically shifts sideways and the hands raise correctly in an asymmetric manner to incorporate turning while jumping on one leg.

\subsection{Style mixing}
\begin{figure}
    \centering
    \includegraphics[width=0.51\textwidth]{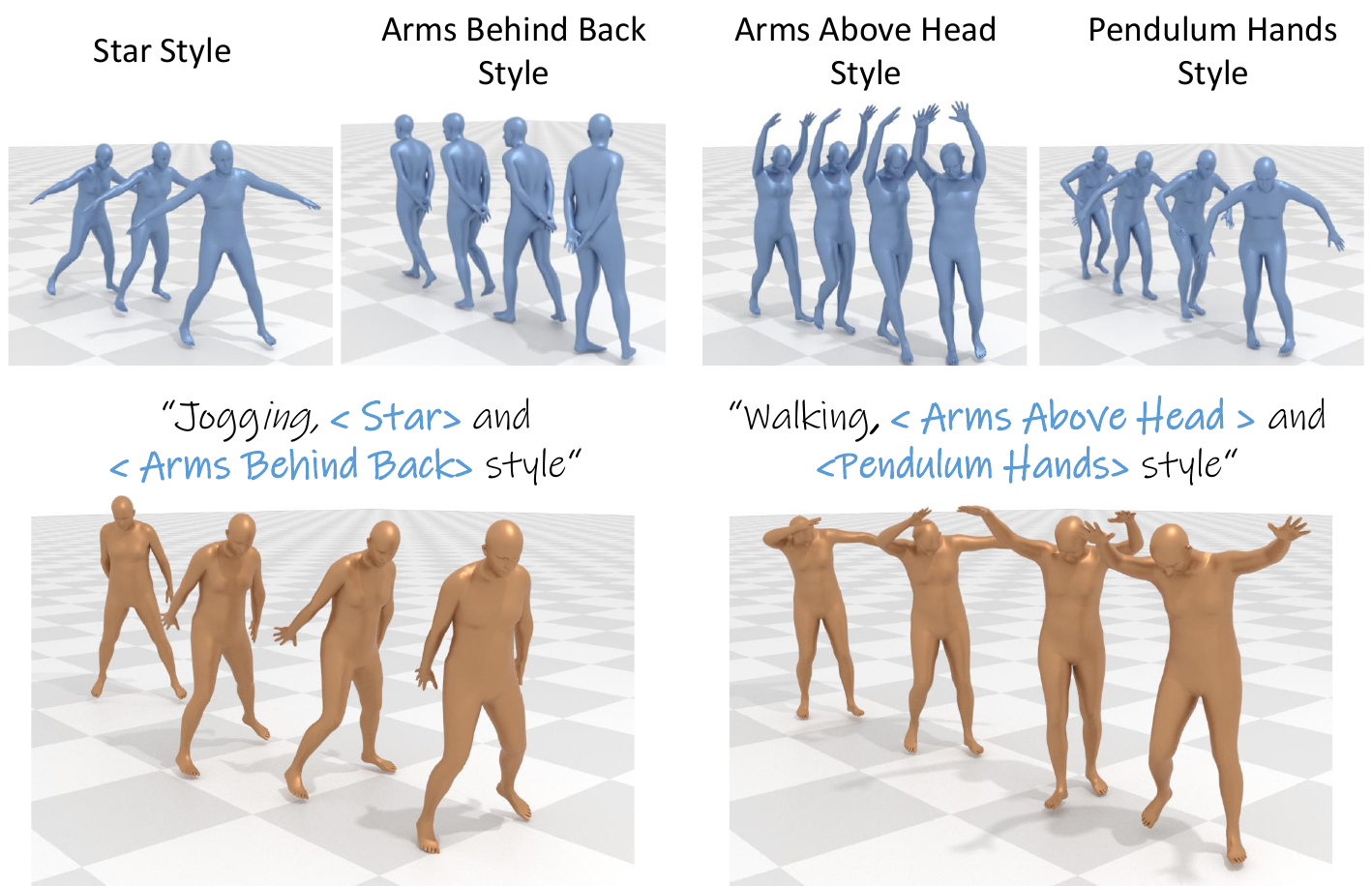}
    \caption{
    \textbf{Style Mixing.} 
    By encoding two styles within the same LoRA using separate tokens, \method generates motions that seamlessly blend both styles.
    \vspace{-15pt}
    }
    \label{fig:mixing}
\end{figure}

Figure~\ref{fig:mixing} presents how \method can be extended to fuse two styles.
Each style $s$ is assigned a unique token $\langle s \rangle$, allowing multiple styles to be incorporated within a single set of LoRA weights by training on them simultaneously.
For each style $s_i$, we initialize a new token $\langle s_i \rangle$ and replace $\langle s \rangle$ with $\langle s_i \rangle$ in $\LL_{style}$ during training.  

By embedding multiple style tokens within the same LoRA module, we enable style mixing - where conditioning on multiple tokens blends their stylistic attributes. 
Although each style is learned independently, combining tokens within a prompt results in a coherent fusion of styles.
For example, a prompt like \textit{"A person is [...] in \textbf{$\langle s_1 \rangle$} style and in \textbf{$\langle s_2 \rangle$} style"} generates a motion that seamlessly integrates both styles.
Notably, not all style pairs can be mixed. For instance, \emph{hands above head} and \emph{arms behind the back} are physically conflicting and cannot be combined.

\section{Discussion}
\label{sec:discussion}

In this paper we introduced \method, a method for motion stylization. 
We demonstrate that in the motion domain, similarly to images, a new concept can be generalized to unseen context through manifold preservation of a pre-trained prior. In essence, we showed that the manifold shift induced by low-rank adaptation outperforms in terms of quality and editability.
Furthermore, we hope this first demonstration of a LoRA employed for motion can guide when and how to use low-rank adaptation in this domain.

Our experimentation also highlights some of the well known problems of style transfer in general, and in the motion domain specifically. Similar to the image domain, the metrics used are unreliable. There is no metric beside the subjective one to measure the quality of an airplane style kicking motion. All automatic evaluators are either sensitive to the action performed and are unfamiliar with most styles, or vice versa. In both cases, stylized actions are out-of-domain and poorly evaluated.
This problem is even more acute for the motion domain, as data is more scarce. There are only a handful of style datasets available, consisting few actions, and demonstrating unclear styles. For example, the largest dataset, which we used, includes styles that are actually actions (e.g., punching) or are hard to define or distinguish (e.g., Star and Monk). Indeed, most literature employs only a subset of the styles for training and evaluation since most styles in the dataset are confusing
\ifarxiv
    (see Appendix~\ref{subsec:sup_data}).
\fi
In the future, we hope this work
helps set common practices for applying LoRA, and facilitate applying it to other motion domains, extending motion generation capabilities in other underrepresented axes except style.

\ifarxiv
\section*{Acknowledgments}
This paper was partially funded by the Israel Science Foundation (1337/22) and Snap Inc.
\fi

{
    \small
    \bibliographystyle{ieeenat_fullname}
    \bibliography{bib}
}

\ifarxiv
    \clearpage
    \appendix
    \section*{\textbf{Appendix}}
    
\section{Implementation Details}
\label{app:impl}

\subsection{Base Model}
Our base MDM~\cite{tevet2023human} model uses an 8-layer transformer decoder with latent dimension $512$, where the text is encoded using DistilBERT~\cite{sanh2019distilbert} and injected through the cross-attention layers. The model was trained with $T=100$ for $500K$ training steps with batch size $64$ on a single A100 GPU. We used the implementation of Tevet et al. \cite{tevet2022motiondiffusion}.

\subsection{LoRA}
We trained a LoRA for all the attention matrices in the model with a rank of $5$, i.e., about 2\% of the parameters of the original matrix. We used $\lambda=1$ and fine-tuned for each style with $4K$ training steps on a single L40S GPU, using a batch containing all motions of the given style and a learning rate of $10^{-5}$. We attached weights to $W^Q, W^K$ and $W^V$ and trained using $T=1000$ diffusion steps, which we found more effective.

\subsection{Style Classifier}

The SRA metric is measured by a pre-trained style classifier.
We find that SMooDi used the same style classifier for both classifier-based guidance and SRA evaluation, which may induce an unwanted or adversarial bias.
Instead, we trained a separate CNN classifier to ensure a fair comparison. 
Our classifier is a two-layer 1D ConvNet with kernel size $3$, followed by temporal average pooling and a single linear layer. The classifier was trained on the 100STYLE dataset, with one motion per style held out as a validation set. We trained with a batch size of $64$ and a learning rate of $5\cdot 10^{-4}$ for $500$ epochs until convergence. The code will be made available.

\section{Datasets}

\subsection{Motion Representation}
\label{subsec:sup_repr}
Following Guo et al.~\cite{Guo_2022_CVPR}, each motion sequence is represented as a time series of 22 joint positions and rotations. Specifically, each pose is encoded as a 263-dimensional vector:  

\[
(\dot{r}^{a}, \dot{r}^x, \dot{r}^z, r^y, j^p, j^r, j^v, c^f) \in \mathbb{R}^{F}.
\]

The first four components describe root motion, where $\dot{r}^{a}$ is the root angular velocity around the Y-axis, $\dot{r}^x$ and $\dot{r}^z$ are the root’s linear velocities in the horizontal plane, and $r^y$ is the root’s vertical position. 
The components $j^p, j^r$, and $j^v$ represent the positions, rotations, and velocities of the joints relative to the root, respectively.
Finally, $c^f$ consists of binary foot contact features, indicating which feet are in contact with the ground for every frame.

\subsection{100STYLE}
The 100STYLE dataset ~\cite{mason2022local} consists of motion capture clips of seven movement types (Backwards Running, Backwards Walking, Forwards Running, Forwards Walking, Idling, Sidestep Running, and Sidestep Walking) performed in 100 different styles.
The dataset was post-processed by Zhong et al.~\cite{zhong2025smoodi}, who primarily retargeted to the SMPL ~\cite{SMPL:2015} skeleton and segmented into short clips, each contains 196 frames.
Some of the released styles were poorly processed and were excluded from all our experiments. See Table~\ref{tab:100style} for details.

\newcolumntype{P}[1]{>{\centering\arraybackslash}p{#1}}

\begin{table*}
\centering
\begin{tabular}{|c|P{10cm}|}
\hline
\textbf{Category} & \textbf{Label} \\ \hline
Character & Aeroplane, Cat, Chicken, Dinosaur, FairySteps, Monk, Morris, Penguin, Quail, Roadrunner, Robot, Rocket, Star, Superman (14) \\ \hline
Personality  & Balance, Heavyset, Old, Rushed, Stiff (5) \\ \hline
Emotion & Angry, Depressed, Elated, Proud (4) \\ \hline
Action & Akimbo, ArmsAboveHead, ArmsBehindBack, ArmsBySide, ArmsFolded, BeatChest, BentForward, BentKnees, BigSteps, BouncyLeft, BouncyRight, CrossOver, FlickLegs, Followed, GracefulArms, HandsBetweenLegs, HandsInPockets, HighKnees, KarateChop, Kick, LeanBack, LeanLeft, LeanRight, LeftHop, LegsApart, LimpLeft, LimpRight, LookUp, Lunge, March, Punch, RaisedLeftArm, RaisedRightArm, RightHop, Skip, SlideFeet, SpinAntiClock, SpinClock, StartStop, Strutting, Sweep, Teapot, Tiptoe, TogetherStep, TwoFootJump, WalkingStickLeft, WalkingStickRight, Waving (53) \\ \hline
Motivation & CrowdAvoidance, InTheDark, LawnMower, OnHeels, OnPhoneLeft, OnPhoneRight, OnToesBentForward, OnToesCrouched, Rushed (9) \\ \hline
Objective & DragLeftLeg, DragRightLeg, DuckFoot, Flapping, ShieldedLeft, ShieldedRight, Swimming, SwingArmsRound, SwingShoulders (9)\\ \hline
Excluded & Zombie, WiggleHips, WhirlArms, WildArms, WildLegs, WideLegs (6)\\ \hline
\end{tabular}
\caption{The detailed grouping of style labels in the 100STYLE dataset, as presented in SMooDi~\cite{zhong2025smoodi} and adopted by this work.}
\label{tab:100style}
\end{table*}

\section{Experiments}

\subsection{Metrics}
Motion metrics, as suggested by Guo et al. \cite{Guo_2022_CVPR}, are based on evaluators comprised of motion encoder $\EE_M$ and text encoder $\EE_T$. Both encoders are trained using contrastive loss to embed the motion and text to a shared space.

\paragraph{SRA.} Top-$k$ accuracy of a trained style classifier, measuring how well the generated motions align with the intended styles.
Since SMooDi employs the same style classifier for both classifier-based guidance and SRA evaluation, we used a separate classifier to ensure a fair comparison-see the details above.  

\paragraph{FID.} 
Fr\'echet Inception Distance (FID) measures the similarity between the distribution of generated motions and real motions. It computes the Fréchet distance between the embeddings of both sets, extracted using the encoder $\EE_M$.  
A lower FID score indicates that the generated motions are more similar to real human motions in terms of distribution, suggesting higher realism and quality.  

\paragraph{MM Dist.} Mean distance of generated motions embeddings and their corresponding texts in a shared embedding space. Lower values indicate that the generated motion adheres to the text condition better.

\paragraph{R-precision.} The percentage of retrieval hits, where a hit is counted if the embedding distance between a generated motion and its corresponding text description is among the top-$k$ closest matches within a batch of 32. Higher values indicate that generated motion adheres to the text condition better.

\paragraph{Foot skating} Measures the percent of adjacent frames where both feet are below a height threshold of 0.05 above the floor in both frames, and the velocity of the feet is greater than 0.5, averaged across all motions. A lower value indicates a higher quality motion.

\paragraph{Diversity.} The mean $L^2$ norm of the distances between randomly sampled motion embeddings within the set. To evaluate diversity preservation, we compute the absolute difference between the diversity of the generated motions and the diversity of the ground truth motions.  
Closer diversity scores indicate that the variability in the generated motions closely matches the diversity observed in real human motions.

\subsection{Data}
\label{subsec:sup_data}
Since certain styles inherently contain content-specific information (e.g., $\langle OnHeels \rangle$), which may interfere with the text prompt, we evaluate the SRA metric across three distinct style groups: (1) all available styles, (2) styles without explicit content-related attributes, excluding the "Action" category in Table~\ref{tab:100style}, as proposed by Zhong et al.\cite{zhong2025smoodi}, and (3) character-based styles (e.g., \textit{Cat, Robot}), corresponding to the "Character" category in Table~\ref{tab:100style}.
Entries marked as "Excluded" were poorly processed and excluded from all our experiments.
Ablation studies were conducted using the first style group, "Characters".

\subsection{Quantitative Results}
Our full quantitative evaluation is presented in Table \ref{tab:quantitative_full}.

\begin{table*}[t!]
    \centering
    \begin{adjustbox}{max width=\textwidth }
        \begin{tabular}{cccccccccccc}
            \toprule
             Style group  & Method  & \makecell{SRA \\ (Top-5)} $\uparrow$ & \makecell{SRA \\ (Top-3)} $\uparrow$ & \makecell{SRA \\ (Top-1)} $\uparrow$ & \makecell{R-precision \\(Top-3)} $\uparrow$ & \makecell{R-precision \\(Top-2)} $\uparrow$ & \makecell{R-precision \\(Top-2)} $\uparrow$ & FID $\downarrow$  & MM Dist $\downarrow$   & Foot skating $\downarrow$   & Diversity  $\rightarrow$  \\
            \hline
            Character & \method & 69.7\% & 56.0\% & 32.1\% & 79.1\%  & 68.4\%  & 48.5\%  & 0.54 & 3.08 & 0.044 & 9.49 \\
            Character & Prompting & 40.6\% & 26.0\% & 8.6\% & 82.5\%  & 73.8\%  & 53.6\%  & 0.33 & 2.86 & 0.046 & 9.73 \\
            Character & SMooDi & 74.9\% & 60.5\% & 31.7\% & 54.6\%  & 43.3\%  & 28.1\%  & 2.94 & 4.68 & 0.102 & 7.61 \\
            Character & MoMo & 98.0\% & 95.7\% & 80.7\% & 12.5\%  & 8.9\%  & 4.7\%  & 16.27 & 8.40 & 0.295 & 7.41 \\
            \hline
            No action & \method & 39.1\% & 31.0\% & 17.6\% & 79.7\%  & 68.4\%  & 48.8\%  & 0.46 & 3.04 & 0.046 & 9.87 \\
            No action & Prompting & 12.8\% & 8.3\% & 2.9\% & 80.1\%  & 70.5\%  & 52.1\%  & 0.31 & 2.96 & 0.045 & 9.64 \\
            No action & SMooDi & 40.4\% & 31.4\% & 16.8\% & 55.9\%  & 45.2\%  & 29.7\%  & 1.70 & 4.49 & 0.124 & 7.77 \\
            No action & MoMo & 83.1\% & 73.3\% & 49.5\% & 12.2\%  & 8.1\%  & 4.2\%  & 18.96 & 8.59 & 0.328 & 6.30 \\
            \hline
            All & \method & 26.5\% & 20.6\% & 11.2\% & 79.0\%  & 68.2\%  & 48.6\%  & 0.50 & 3.09 & 0.042 & 8.58 \\
            All & Prompting & 8.3\% & 5.2\% & 2.0\% & 79.7\%  & 70.5\%  & 52.2\%  & 0.30 & 3.06 & 0.045 & 10.02 \\
            All & SMooDi & 27.5\% & 20.2\% & 10.1\% & 57.3\%  & 45.9\%  & 30.7\%  & 1.37 & 4.52 & 0.115 & 9.68 \\
            All & MoMo & 68.7\% & 59.6\% & 37.7\% & 11.7\%  & 8.2\%  & 4.2\%  & 19.52 & 8.67 & 0.315 & 6.41  \\ 
            \bottomrule
        \end{tabular}
    \end{adjustbox}
    \caption{
    \textbf{Full Quantitative results.}
    }
    \label{tab:quantitative_full}
\end{table*}

\subsection{Ablations Study}
\label{subsec:sup_ablations}
Our full ablation study is presented in Table \ref{tab:ablations_full}

\begin{table*}[t!]
    \centering
    \begin{adjustbox}{max width=\textwidth }
        \begin{tabular}{lccccccccccc}
            \toprule
              & \makecell{SRA \\ (Top-5)} $\uparrow$ & \makecell{SRA \\ (Top-3)} $\uparrow$ & \makecell{SRA \\ (Top-1)} $\uparrow$ & \makecell{R-precision \\(Top-3)} $\uparrow$ & \makecell{R-precision \\(Top-2)} $\uparrow$ & \makecell{R-precision \\(Top-2)} $\uparrow$ & FID $\downarrow$  & MM Dist $\downarrow$   & Foot skating $\downarrow$   & Diversity  $\rightarrow$  \\
            \hline
$rank=1$ & 57.3\% & 40.2\% & 16.6\% & 82.1\%  & 71.7\%  & 51.5\%  & 0.29 & 2.90 & 0.037 & 8.74 \\
 $rank=3$ & 62.1\% & 47.9\% & 24.3\% & 81.4\%  & 71.1\%  & 50.6\%  & 0.46 & 3.00 & 0.039 & 9.36 \\
 $rank=5$ & 70.8\% & 57.7\% & 33.3\% & 78.8\%  & 68.1\%  & 48.9\%  & 0.54 & 3.09 & 0.042 & 10.15 \\
 $rank=10$ & 79.8\% & 69.3\% & 44.8\% & 77.5\%  & 66.2\%  & 46.2\%  & 0.94 & 3.25 & 0.046 & 9.13 \\
 $rank=20$ & 84.5\% & 76.0\% & 52.6\% & 75.2\%  & 64.6\%  & 44.4\%  & 1.13 & 3.38 & 0.050 & 8.93 \\
 $rank=4$ & 65.3\% & 50.5\% & 25.9\% & 79.4\%  & 68.8\%  & 48.4\%  & 0.45 & 3.02 & 0.038 & 7.71 \\
 $rank=6$ & 73.5\% & 61.4\% & 36.7\% & 78.3\%  & 66.9\%  & 47.2\%  & 0.64 & 3.10 & 0.043 & 8.95 \\
 $rank=7$ & 75.7\% & 62.5\% & 38.3\% & 77.7\%  & 67.0\%  & 47.2\%  & 0.68 & 3.18 & 0.043 & 8.74 \\
 $rank=15$ & 82.4\% & 72.8\% & 48.1\% & 75.1\%  & 63.9\%  & 43.8\%  & 1.03 & 3.29 & 0.048 & 9.31\\
            \hline
             $\lambda=0$ & 85.4\% & 76.8\% & 52.2\% & 75.4\%  & 64.4\%  & 44.6\%  & 0.72 & 3.32 & 0.075 & 9.58 \\
 $\lambda=0.1$ & 78.2\% & 66.3\% & 40.3\% & 79.4\%  & 68.4\%  & 48.5\%  & 0.34 & 3.07 & 0.064 & 10.03 \\
 $\lambda=0.25$ & 73.8\% & 60.8\% & 35.4\% & 79.6\%  & 68.9\%  & 49.4\%  & 0.31 & 3.02 & 0.054 & 10.30 \\
 $\lambda=0.5$ & 71.6\% & 58.7\% & 33.1\% & 79.4\%  & 69.0\%  & 49.8\%  & 0.38 & 3.04 & 0.047 & 10.26 \\
 $\lambda=0.75$ & 70.7\% & 57.8\% & 32.8\% & 79.2\%  & 67.9\%  & 49.6\%  & 0.46 & 3.07 & 0.044 & 10.19 \\
 $\lambda=0.85$ & 70.6\% & 57.5\% & 33.1\% & 79.2\%  & 68.0\%  & 49.0\%  & 0.49 & 3.08 & 0.043 & 10.19 \\
 $\lambda=1$ & 70.6\% & 57.7\% & 33.3\% & 78.8\%  & 67.9\%  & 48.7\%  & 0.55 & 3.09 & 0.042 & 10.13 \\
 $\lambda=2.5$ & 67.0\% & 52.5\% & 28.5\% & 79.3\%  & 68.3\%  & 48.1\%  & 0.69 & 3.10 & 0.039 & 10.22 \\
 $\lambda=5$ & 53.8\% & 38.7\% & 16.6\% & 79.7\%  & 69.5\%  & 48.9\%  & 0.56 & 3.02 & 0.037 & 10.33 \\
            \hline
             Generated & 79.4\% & 66.6\% & 42.8\% & 63.1\%  & 52.5\%  & 34.8\%  & 2.33 & 4.15 & 0.057 & 8.66 \\
 HumanML3D & 70.7\% & 57.6\% & 33.0\% & 78.9\%  & 67.9\%  & 48.8\%  & 0.54 & 3.09 & 0.042 & 10.14 \\
 Mixed & 76.4\% & 63.8\% & 38.4\% & 75.6\%  & 63.7\%  & 45.3\%  & 1.24 & 3.37 & 0.034 & 9.60 \\
 No $\LL_{prior}$ & 85.4\% & 76.8\% & 52.2\% & 75.4\%  & 64.4\%  & 44.6\%  & 0.72 & 3.32 & 0.075 & 9.58 \\
            \hline

 \textit{w $W^Q$, w/o FFN} & 70.8\% & 57.7\% & 33.4\% & 79.0\%  & 68.0\%  & 48.8\%  & 0.54 & 3.09 & 0.042 & 10.12 \\
 \textit{w $W^Q$, w FFN}  & 95.5\% & 93.6\% & 80.0\% & 16.7\%  & 11.6\%  & 6.1\%  & 9.61 & 7.49 & 0.087 & 7.08 \\
 \textit{w/o $W^Q$, w/o FFN} & 69.2\% & 56.4\% & 32.7\% & 78.6\%  & 68.5\%  & 49.0\%  & 0.49 & 3.05 & 0.041 & 9.55 \\
 \textit{w/o $W^Q$, w FFN} & 93.5\% & 90.1\% & 76.0\% & 17.9\%  & 12.7\%  & 7.0\%  & 9.33 & 7.31 & 0.082 & 6.13 \\

            \bottomrule
        \end{tabular}
    \end{adjustbox}
    \caption{
    \textbf{Full ablation study.}
    }
    \label{tab:ablations_full}
\end{table*}

\fi

\end{document}